\documentclass[runningheads]{llncs}
\usepackage{lineno}
\usepackage[T1]{fontenc}
\usepackage{microtype}
\usepackage{graphicx}
\usepackage{url}
\usepackage{booktabs}   
\usepackage{multirow}   
\usepackage{longtable}
\usepackage{xcolor}
\usepackage{academicons}
\usepackage[hidelinks]{hyperref}

\newcommand{\repeatthanks}{\textsuperscript{\thefootnote}}
\makeatletter
\def\@fnsymbol#1{\ensuremath{\ifcase#1\or\star\or\dagger\or{\star\star\star}\or\ddagger\or
    \mathchar "278\or \mathchar "27B\or \|\or **\or \dagger\dagger
\or \ddagger\ddagger \else\@ctrerr\fi}}
\makeatother
\begin{document}

\title{Rethinking the Role of Feature Engineering and Learning Strategies in Few-Shot Hidden Emotion Recognition}
\titlerunning{ReFELS}

\author{
  Xiaochuan Guo\thanks{Equal contribution.}\inst{1}\orcidID{0009-0007-8105-3719} \and
  Jihao Gu\repeatthanks\inst{2}\orcidID{0009-0009-0141-4807} \and
  Haixu Liu\repeatthanks\inst{3,4}\orcidID{0009-0007-8115-0826} \and
  Yuxin Liu\repeatthanks\inst{5}\orcidID{0009-0003-9995-9173} \and
  Qi Wang\inst{6}\orcidID{0009-0003-8390-3896} \and
  Yufei Wang\inst{7}\orcidID{0009-0008-6002-3729} \and
  Fei Wang\thanks{Corresponding authors: Fei Wang and Kun Li.
  Emails: \texttt{jiafei127@gmail.com}, \texttt{kunli.hfut@gmail.com}.}\inst{1}\orcidID{0009-0004-1142-6434}
  \and
  Kun Li\repeatthanks\inst{8}\orcidID{0000-0001-5083-2145}
  \and
  Dan Guo\inst{1,5}\orcidID{0000-0003-2594-254X}
}

\authorrunning{X. Guo, J. Gu, H. Liu et al.}

\institute{
  Hefei University of Technology, China
  \and University College London, United Kingdom
  \and The University of Sydney, Australia
  \and Beijing QBoson Quantum Technology Co., Ltd., China
  \and Institute of Artificial Intelligence, Hefei  Comprehensive \\ National Science Center, China
  \and Beihang University, China
  \and The University of New South Wales, Australia
  \and United Arab Emirates University, United Arab Emirates
}

\maketitle

\begin{abstract}
  In this paper, we present the solution developed by our team, XInsight Lab, which achieved first place in Track 3 of the 4th EI-MIGA-IJCAI Challenge with the test accuracy of 0.76923. To address the challenge of weak and sparse implicit emotion evidence in long videos, this paper extends the winning solution from the previous competition and proposes a compact multi-modal temporal modeling framework.
  The framework integrates and evaluates the effects of multi-source features, including 2D/3D skeletons, facial expression Blendshapes, DINOv2/v3 vision foundation models, X-CLIP video features, and Gemini semantic priors. Architecturally, we propose a cross-attention mechanism that utilizes static pose features (Base) as the Query and dynamic micro-motion differential features (Offset) as the Key/Value. By capturing local relative velocities, this mechanism eliminates static biases related to individual body shape and identity. Concurrently, an adaptive pooling method based on Multiple Instance Learning (MIL) is employed to extract instantaneous emotions while suppressing background noise in long sequences.
  Finally, the paper reveals the "representation collapse" phenomenon of general vision foundation models in micro-dynamic tasks, and analyzes the underlying mechanisms where networks fall into public-leaderboard-driven "pseudo-generalization" due to shortcut learning and rote memorization.

\end{abstract}


\section{Introduction}

\subsection{Background}

Behavior-based hidden emotion understanding aims to infer latent affective states from subtle, spontaneous, and identity-insensitive nonverbal behaviors~\cite{liu2021imigue,chen2023smg,wang2026imigue,gu2025motion}. Different from conventional facial expression recognition or action recognition, the discriminative evidence is often weak, sparse, and distributed over long videos.
In the iMiGUE setting~\cite{liu2021imigue}, videos are muted post-match tennis interviews, and emotional polarity is inferred from the match outcome. This protocol reduces leakage from speech but makes the task strongly dependent on visual and behavioral cues. It also introduces three practical challenges: limited labeled data, class imbalance, and short-lived affective evidence that may only appear in a few frames.

Recent SOTA systems suggest that robust hidden emotion understanding requires complementary representations. Human-centric landmarks suppress background interference, self-supervised visual features encode appearance and context, and VLM-generated semantic descriptions provide high-level behavioral priors~\cite{oquab2023dinov2,simeoni2025dinov3,team2023gemini,wang2024eulermormer,wang2024frequency}. Prior challenge solutions further show the effectiveness of person cropping, whole-body pose features, temporal offsets, pseudo labels, and modality-wise training~\cite{wang2026weak,gu2025mmgesture}.
However, these methods leave an important modeling question underexplored: In the extreme setting of multimodal sparse features and few-shot supervision, what kind of feature engineering architecture and temporal learning strategy can genuinely drive the model to solidify robust representations with strong generalizability?
Furthermore, within the weak-supervision iteration and voting ensemble paradigm relying on Vision-Language Models (VLMs), does the performance leap of downstream models stem from genuinely acquiring deep domain knowledge, or merely from memorizing noisy pseudo-labels, thereby achieving an empirical "pseudo-generalization" amid the prior randomness of the ensembled foundation models?

We address this question with a compact multimodal temporal framework. Our input contains body, face, and hand landmarks, facial expression descriptors, OpenPose whole-body keypoints, DINOv2/DINOv3 appearance features, X-CLIP video features, Gemini-based textual analyses encoded by BERT, optical-flow descriptors, and Depth Anything 3 (DA3) depth features~\cite{lugaresi2019mediapipe,baltrusaitis2018openface,cao2021openpose,oquab2023dinov2,simeoni2025dinov3,ni2022xclip,team2023gemini,devlin2019bert,raft,depthanything3}. For each temporal stream, we construct frame-wise offset features to capture small displacements that are often more discriminative than static coordinates or frame-level appearance alone, while optical flow and DA3 provide complementary motion and geometry cues.

Architecturally, to further investigate how base-offset features contribute to the model's overall performance, we explore two distinct fusion designs and a two-stage protocol. The first, \emph{Independent}, regards original streams and offset streams as separate modalities and learns one temporal encoder per stream. The second, \emph{Base-Offset Cross-Attention}, synchronously samples the base and offset sequences and uses base features as queries while using offsets as keys and values, following the Transformer attention formulation~\cite{vaswani2017attention}. Each stream is encoded by Transformer blocks with positional encoding, temporal consistency modeling, and attention pooling. The final multimodal representation is obtained by late concatenation and optimized with a lightweight MLP classifier.
Meanwhile, we adopt a two-stage protocol: modality-wise pretraining followed by joint fine-tuning, augmented by iterative pseudo-labeling via voting ensembles for semi-supervised learning.
\subsection{Contributions}
Our contributions are threefold.
\begin{itemize}
  \item \textbf{We propose} a novel perspective for hidden emotion recognition by reformulating it as offset-aware multimodal temporal modeling rather than generic video classification, which effectively guides the network to capture weak and sparse behavioral evidence.
  \item \textbf{We design} a rigorous controlled comparison between independent motion encoding and explicit base-offset cross-attention under identical feature and training protocols. Through this, we deeply investigate the underlying logic of performance gains and systematically expose critical pitfalls of ``false improvements.''
  \item \textbf{We demonstrate} the exceptional efficacy and superiority of our framework by achieving a leaderboard accuracy of 0.76923, ranking first in Track 3 of the 4th EI-MiGA-IJCAI Challenge.
\end{itemize}

\section{Related Work}

\paragraph{\textbf{Data Preprocessing and Human-centric Representations.}}
Effective preprocessing is central to hidden emotion understanding because useful cues are often local, brief, and easily overwhelmed by background or camera variation. A common strategy is to convert raw videos into human-centric representations~\cite{li2023data,wang2026gait,wang2025exploiting,liu2026self,shen2026spatial}. OpenPose~\cite{cao2021openpose} estimates whole-body keypoints with body, face, and hand landmarks using part affinity fields, while MediaPipe~\cite{lugaresi2019mediapipe} provides efficient perception pipelines for body and hand tracking. Facial behavior toolkits such as OpenFace~\cite{baltrusaitis2018openface} further extract facial landmarks, head pose, gaze, and action-unit or expression descriptors. More recent whole-body pose estimators, including RTMPose, DWPose, ViTPose, and Sapiens, improve the speed, accuracy, or resolution of pose extraction through coordinate classification, two-stage distillation, plain vision Transformers, and large-scale human-centric pretraining~\cite{jiang2023rtmpose,yang2023dwpose,xu2022vitpose,khirodkar2024sapiens}. These pose features reduce background noise but remain sensitive to occlusion, low-resolution hands, and head-pose changes.

Beyond coordinates, appearance and semantic features provide complementary evidence. Self-supervised image encoders such as DINOv2 and DINOv3 produce robust visual representations without task-specific fine-tuning~\cite{gu2025performance,oquab2023dinov2,simeoni2025dinov3,wang2026task}, and video-language models such as X-CLIP extend image-language pretraining to temporal recognition~\cite{ni2022xclip}. VLMs can further generate structured behavioral descriptions that are encoded by language models such as BERT~\cite{team2023gemini,Gu_2026_CVPR,devlin2019bert}. Another line of preprocessing reconstructs 3D facial or avatar geometry, e.g., FLAME-based faces and 3D Gaussian avatars, to obtain view-consistent deformation cues~\cite{li2017flame,kerbl2023gaussian}. We adopt a lighter scalable alternative by extracting DA3-based monocular depth descriptors, which provide geometry-aware cues without explicit avatar reconstruction~\cite{depthanything3}. Optical flow is also used to encode short-term pixel-level motion, complementing landmark offsets that only describe motion on detected human keypoints~\cite{raft}. In this work, we focus on a scalable 2D/appearance/motion/depth/semantic feature setting and explicitly augment temporal streams with frame-wise offsets when applicable.

\paragraph{\textbf{Models for Micro-gesture and Hidden Emotion Understanding.}}
Early video-based emotion or behavior recognition methods mainly rely on RGB backbones and generic temporal aggregation.
I3D, TSM, and Video Swin Transformer model spatiotemporal appearance patterns from clips or frame sequences~\cite{carreira2017quo,lin2019tsm,liu2022video}. Skeleton-based methods such as ST-GCN and PoseConv3D replace pixels with structured keypoints and exploit spatiotemporal modeling over human joints~\cite{yan2018stgcn,duan2022revisiting}. While these architectures are effective for action recognition, hidden emotion understanding is harder because the relevant motion is subtle, non-periodic, and often not aligned with the anatomical adjacency used by graph models.
Recent micro-gesture systems therefore move toward multi-stream modeling. RGB, skeleton, optical flow, depth, joint, limb, and textual streams have been combined through late fusion, cross-modal attention, prototype refinement, or modality-weighted ensembling~\cite{li2023joint,chen2024prototype,huang2024multiscale,gu2025mmgesture,liu2024micro,liu2025online,wang2025robust}.
Weak-to-Strong further shows that VLM-based pseudo labels, DINOv2 appearance features, and OpenPose offsets can improve multimodal hidden emotion recognition under limited supervision~\cite{wang2026weak}.
Regarding training strategies, existing systems often adopt temporal sampling, modality-wise pretraining, pseudo-label augmentation, focal or reweighted losses, and threshold or ratio calibration to achieve stable evaluation~\cite{lin2017focal,cui2019classbalanced,wang2026weak}.

\paragraph{\textbf{Benchmarks and Dataset Protocols.}}
Several datasets have shaped the study of subtle behavioral cues. SMG~\cite{chen2023smg} provides spontaneous micro-gesture videos for emotional stress state analysis. iMiGUE~\cite{liu2021imigue} focuses on identity-free micro-gesture understanding and emotion analysis from tennis interviews.
More recently, iMiGUE-3K~\cite{wang2026imigue} scales micro-gesture analysis to thousands of long interview videos and supports foundation-model-style evaluation.
Additionally, MA-52~\cite{guo2024benchmarking}, MMA-52~\cite{li2025mmad}, and MA-Bench~\cite{li2026bench} belong to the broader field of subtle behavior analysis~\cite{li2025prototypical,gu2025motion}, their distinct target tasks make them complementary benchmarks for evaluating cross-task generalization. Their use in competitions~\cite{guo2024mac,li2025mac} has also fostered broader community participation and helped drive progress in this field.


\section{Methodology}

\subsection{Deep Feature Engineering and Representation Construction}
To maximize the signal-to-noise ratio (SNR) of input video signals and explicitly expose extremely subtle motion cues within a high-dimensional feature space, this paper constructs a dimensionally rigorous and physically well-defined feature decoupling pipeline tailored for audiovisual signals in complex iMiGUE scenarios.

\begin{figure}[htbp]
  \centering
  \makebox[\linewidth][c]{
    \includegraphics[width=0.935\linewidth]{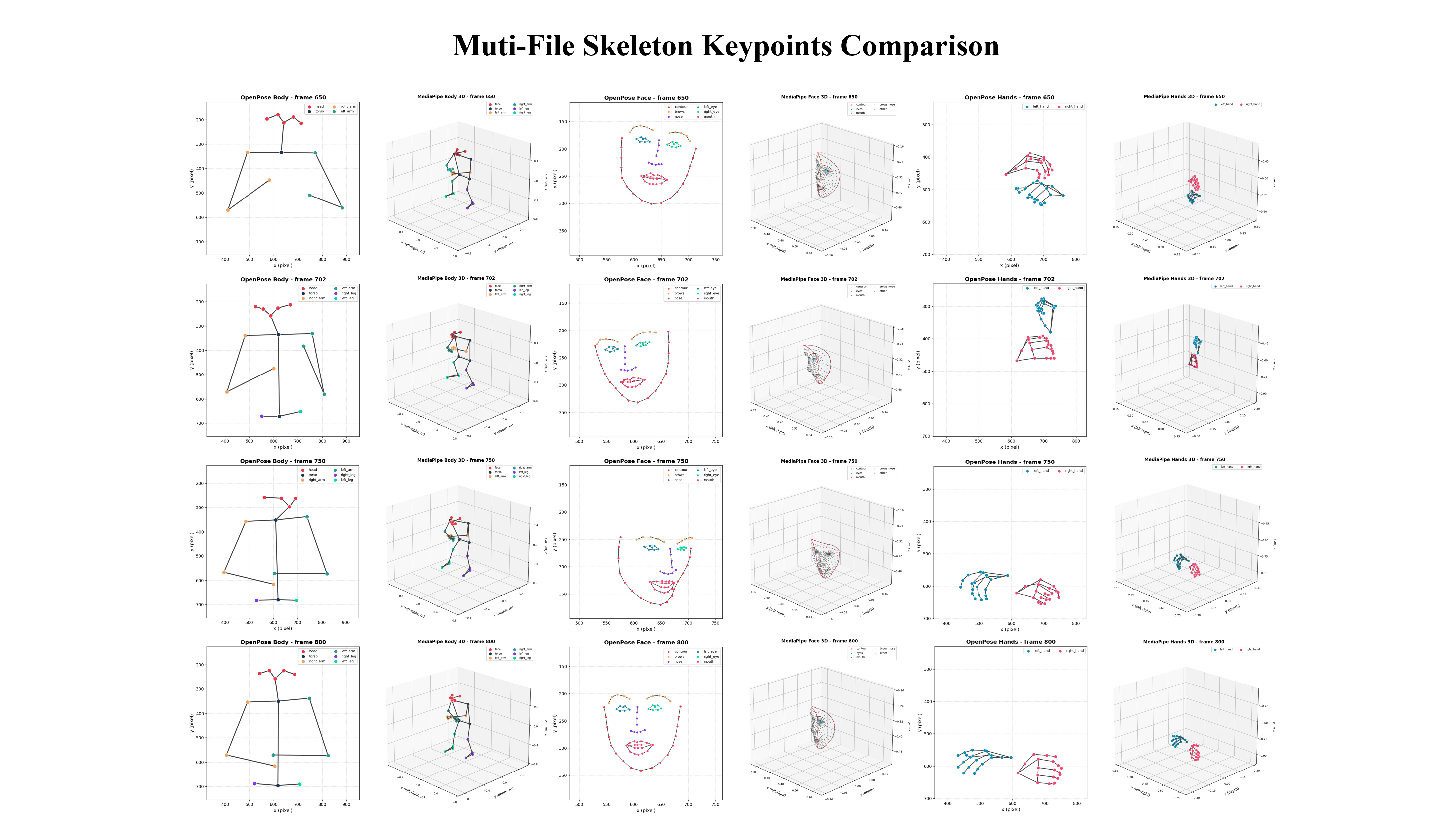}
  }
  \caption{Comparison of keypoint detection between MediaPipe and OpenPose.}
  \label{fig:sample}
\end{figure}

At the level of kinematic and dynamic feature extraction, we perform strict physical differentiation and parallel extraction of human limb and facial micro-movements across both the 2D pixel plane and 3D physical space. For 2D plane features, the OpenPose framework is utilized to densely extract absolute kinematic features based on the image coordinate system, which are then finely deconstructed into a 75-dimensional sequence of body landmarks, a 210-dimensional sequence of face landmarks, and a 126-dimensional sequence of hand landmarks. To capture 3D topologies with deep spatial structures, we concurrently employ the MediaPipe framework to extract 3D spatial coordinates containing depth priors, strictly reconstructing them into 198-dimensional 3D body landmarks, a high-fidelity 1434-dimensional dense 3D facial mesh vertex sequence, and 126-dimensional 3D hand landmarks. As shown in Fig.~\ref{fig:sample}, we provide a visual example of 2D OpenPose keypoints and 3D MediaPipe-based spatial keypoint representations. Furthermore, to accurately and directly quantify localized facial muscle spasms and contractions, we independently derive 52-dimensional continuous facial expression blendshape parameters. To enable downstream networks to explicitly perceive the dynamic evolution of micro-expressions, for all the aforementioned 2D/3D skeleton and expression sequences, we retain the original parameters as ``Base features'' to represent static poses. Concurrently, we compute frame-by-frame numerical differentiation between adjacent valid frames along the temporal axis, explicitly constructing ``Offset features'' of identical dimensions. This operation essentially extracts the first-order derivative of human kinematics at the input level, forcing the model to shift its attention from absolute spatial poses---which suffer from static identity and body-shape biases---to localized relative micro-movement velocities.
At the visual appearance and macro-contextual feature level, to counteract the catastrophic background noise in interview scenarios, we deploy the YOLOv11 object detection algorithm to perform fine, frame-by-frame cropping of the human subject. Subsequently, the purified human image sequences, stripped of background interference, are fed into DINOv2 to extract 1024-dimensional high-fidelity frame-level visual features. Concurrently, we employ an improved YOLOv26 combined with DINOv3 to extract a second set of fine-grained visual representations up to 4096 dimensions. This bounding-box-based spatial hard attention mechanism thoroughly prevents the network from overfitting to irrelevant backgrounds, while the dense receptive fields of the vision foundation models drastically enhance the network's generalization sensitivity toward localized subtle skin movement textures. Moreover, to compensate for the macro-scene contextual semantics lost during video cropping, we utilize a pre-trained X-CLIP model to extract 512-dimensional clip-level global semantic features from the uncropped original video, ultimately forming a complementary visual representation of ``microscopic pure textures and macroscopic global context.''

At the semantic prior level, purely data-driven deep vision networks often lack the psychological common sense required to infer human deceptive or masking behaviors. To address this, we introduce a Vision-Language Large Model (Gemini) to perform high-level semantic cognitive reasoning on video frame sequences. By combining advanced prompting strategies of ``Chain-of-Thought and Reflection'' (CoT+Reflection), we compel the large model to objectively extract Facial Action Coding System Action Units (FACS AUs), followed by logical deduction and self-reflection based on injected psychological prior rules. The generated structured analysis text is subsequently mapped into a 768-dimensional dense semantic feature tensor via the Qwen Embedding text encoder, thereby reducing the dimensionality of invaluable human cognitive logic into continuous variables that can be directly fused by downstream joint networks. Additionally, this paper explores the extraction of deformation offsets of dynamic 3D Gaussian point clouds in a canonical pose space based on the HRAvatar reconstruction algorithm as exploratory geometric topological features. However, extensive empirical validation reveals that certain samples in in-the-wild datasets exhibit extreme rapid camera movements and severe physical occlusions, which easily trigger topological divergence in the underlying 3D reconstruction. To guarantee the absolute inferential robustness of the multimodal engineering system in unknown real-world environments, this exploratory geometric branch was proactively discarded in the final ensemble model.

\subsection{Temporal Aggregation and Multimodal Network Architecture}
Faced with the dozens of heterogeneous temporal feature sequences spanning thousands of frames extracted above, traditional global average pooling would mathematically flatten the abrupt emotional leakage signals occurring across just a few frames into the thousands of neutral background frames in terms of mathematical expectation. To circumvent this, we comprehensively adapt an adaptive attention pooling layer based on Multiple Instance Learning (MIL) at the end of all single-branch temporal encoding modules. This module dynamically computes frame-level attention weights across the entire sequence by autonomously learning a global query vector, endowing the network with highly adaptive temporal retrieval capabilities. This mechanism drives the model to accurately allocate extremely high response weights to peak frames with abrupt emotional changes, achieving lossless compression of effective information while aggressively suppressing long-sequence background noise.

Regarding the cross-modal spatio-temporal interaction topology of multidimensional features, this paper designs two parallel network paradigms for ensemble integration.

\noindent\textbf{Independent Feature Stream Modeling:} Adhering to the philosophy of extreme feature isolation, we treat all 2D and 3D base features, along with their corresponding offset differential features, as completely independent input sources that do not interfere with one another. They are fed separately into dedicated Transformer encoding networks and MIL pooling layers, and are only subjected to high-dimensional feature concatenation and joint semantic alignment at the fully connected layer at the deepest end of the network. This late fusion strategy maximizes the decoupling of heterogeneous features, effectively preventing highly sparse skeleton displacement coordinates and highly dense visual tensors from triggering catastrophic manifold interference in shallow networks.

\begin{figure}[htbp]
  \centering
  \includegraphics[width=\linewidth]{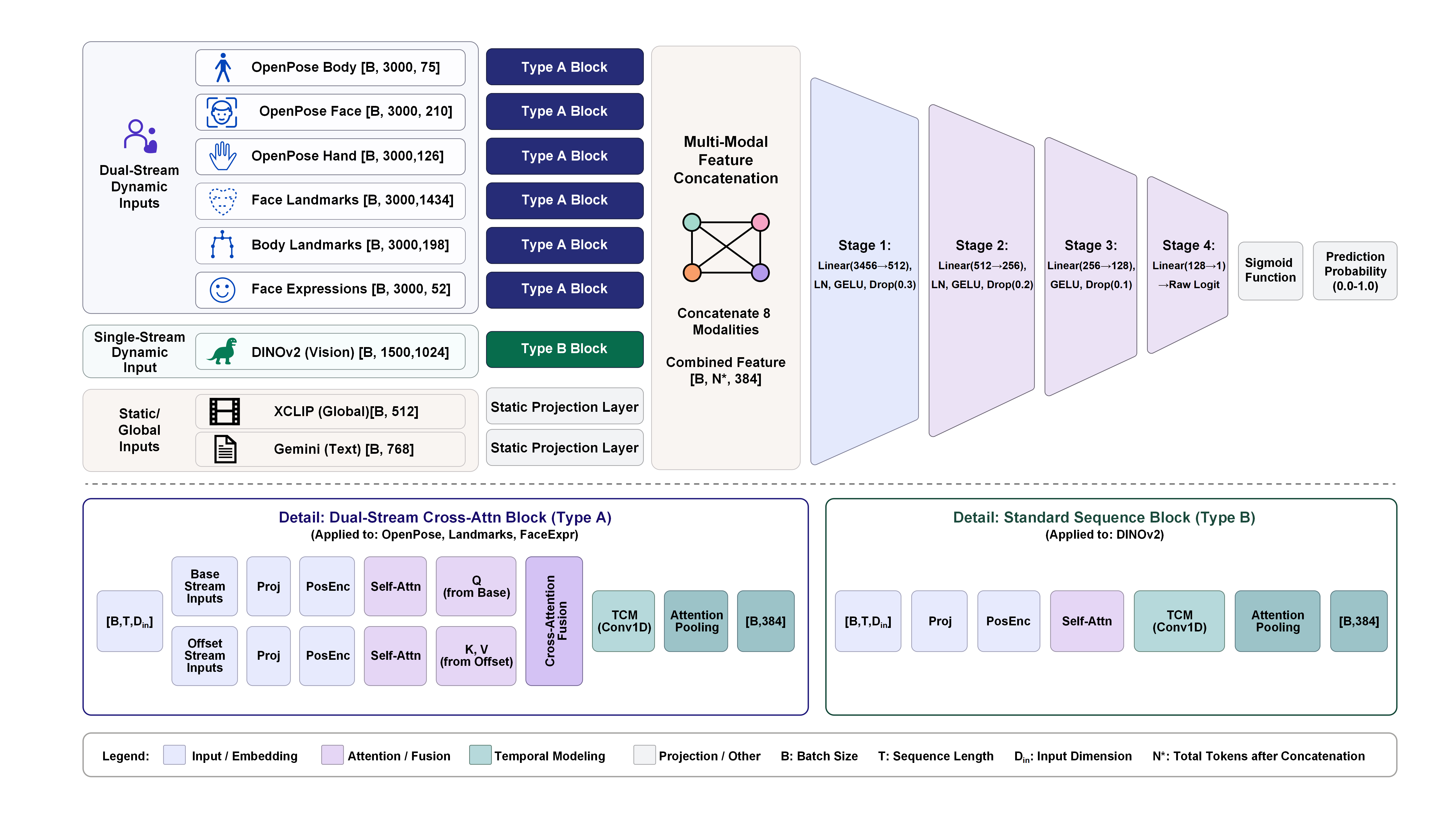}
  \caption{Base-Offset Cross-Attention Model Structure.}
  \label{fig:framework}
\end{figure}

\noindent\textbf{Base-Offset Cross-Attention Modeling:} As illustrated in Fig.~\ref{fig:framework}, for modalities with strict kinematic derivation relationships (such as the 1434-dimensional 3D absolute facial coordinates and their corresponding 1434-dimensional temporal differences), the base features representing static underlying physical constraints serve as the Query vector, while the offset features representing dynamic minute changes serve as the Key and Value vectors ($Key, Value$). The two components complete dense cross-stream interactions within a cross-attention module. This network architecture explicitly injects kinematic synergistic priors at the topological level, empowering the model with high-level reasoning capabilities to dynamically and conditionally evaluate the significance of localized minute deformations under specific underlying physical constraints of the torso and face.
\subsection{Progressive Weakly-Supervised and Iterative Training Strategy}

To address the extreme scarcity of effective strong annotations and the naturally occurring long-tailed distribution imbalance in the iMiGUE dataset, this study employs Focal Loss at the global joint classifier terminal to dynamically adjust the gradient penalty weights of samples. In the specific hyperparameter configuration, we set the initial adjustment factors to $\alpha=0.35$ and $\gamma=3.0$; as the semi-supervised learning-based label iterations progress, $\gamma$ is gradually decreased, and $\alpha$ is moderately adjusted. Distinct from conventional algorithms that excessively inflate the $\gamma$ value to force the model to obsess over hard examples, the core motivation behind our parameter setting is that, in in-the-wild surveillance datasets of concealed emotions, the ``hard examples'' in the model's eyes are often absolute noise and defective footage caused by severe camera defocusing, physical occlusions, or unavoidable subjective human annotation errors. This loss function parameter configuration explicitly prevents the model from over-focusing on invalid, extreme hard examples with exceptionally high losses. From a fundamental perspective, this substantially prevents the deep network from catastrophically memorizing and overfitting to anomalous noise under small-sample conditions, thereby substantially expanding the true generalization boundaries of the classification hyperplane on unseen test domains.
To further optimize the model's predictive performance, we introduced a post-processing strategy based on a positive sample ratio threshold.
By incorporating dynamic feedback from the leaderboard, parameter tuning was conducted within the range of 0.56–0.66 to determine the optimal threshold. The complete reproduction of the entire training pipeline relies on a single NVIDIA RTX Pro 6000 GPU, with a total computation time of approximately 3 days.

Based on the above settings, this paper executes a progressive, three-stage learning pipeline from shallow to deep:

\noindent\textbf{Stage 1: Modality-Specific Independent Pre-training.} At the beginning of model initialization, the system temporarily disconnects all high-level multimodal fusion channels. It relies solely on the limited training set samples with official ground-truth labels to conduct completely independent, purely supervised classification pre-training for each of the aforementioned individual 2D/3D skeleton channels and high-dimensional visual feature extraction branches. This strategy fundamentally circumvents the ``modality inertia and competitive dominance'' pitfalls during the joint initialization of complex multimodal networks, preventing the network from over-relying on the fast-converging dominant visual branches to rapidly minimize the overall loss, which would leave weak feature streams, such as motion differences, in long-term gradient starvation. Independent pre-training forces each feature channel to squeeze its discriminative potential to the absolute limit within its dedicated feature subspace, eliminating the weak-link vulnerability of late fusion at its source.

\noindent\textbf{Stage 2: Weakly-Supervised Joint Fine-Tuning with Priors.} Subsequently, we assemble the optimal network weights from each independent branch in Stage 1 to establish the multimodal fusion network. At this point, we directly extract the zero-human-cost inference predictions generated by the Vision-Language Large Model (Gemini) on the unlabeled test set videos during the feature engineering phase, treating them as initial pseudo-labels. By mixing the ground-truth training set with the pseudo-labeled test set to form an extensive weakly-supervised data pool, the network initiates global cross-domain joint alignment fine-tuning with an extremely conservative and small learning rate. This process not only expands the scale of available data but also compels the high-dimensional classification plane of the joint network to pre-fit and adapt to the vast and unknown underlying data distribution of the test set.

\noindent\textbf{Stage 3: Iterative Self-Training via Multi-Model Voting Ensemble.} As the multimodal joint network progressively converges on specific domain scenarios, its high-level discriminative capability in vertical emotion anti-spoofing tasks is further enhanced. To reduce noise in the initial pseudo-labels and elevate label quality, an iterative self-training mechanism is introduced in this stage. Rather than discarding the initial predictions generated by Gemini, we treat them as prior baselines, combining them with the topologies of multiple currently converged fine-tuned models to form an ensemble voting system. Subsequently, this batch of high-quality pseudo-labels is utilized to re-drive the independent modality training and joint fine-tuning of the previous two stages. The aforementioned process executes iteratively for several rounds, ultimately steering the decision boundary of the joint network toward stability and optimality.


\section{Experiment}
\subsection{Ablation study}
This work extends last year's winning solution by introducing key enhancements. Through systematic ablation experiments, we identified effective training strategies and feature engineering methods tailored for foundation models. As shown in Table~\ref{tab:ablation}, each proposed component consistently improves performance over the baseline, with Multiple Instance Learning (MIL) contributing the most significant gain. Crucially, our results confirm that integrating Multi-Instance Learning (MIL) yields significant performance gains.

\begin{table}[htbp]
  \centering
  \caption{Ablation study of general enhancement strategies based on base-model}
  \label{tab:ablation}
  \renewcommand{\arraystretch}{1.2}
  \begin{tabular}{lcc}
    \toprule
    \textbf{Method} & \textbf{ACC (\%)} & \textbf{$\Delta$ (\%)} \\
    \midrule
    MLP                & 63.46 & --     \\
    + Weak supervision  & 66.35 & +2.89  \\
    + Offset features  & 68.27 & +1.92  \\
    + Pre-training     & 69.23 & +0.96  \\
    + Multiple Instance Learning              & 70.19 & +0.96  \\
    \bottomrule
  \end{tabular}
\end{table}

\subsection{Contrast experiment}
This paper evaluates the performance of independent modality training versus the joint training of dual-architecture models under both supervised and semi-supervised learning frameworks.
The experimental results align with expectations. First, the 3D-based MediaPipe demonstrates superior keypoint representation capabilities compared to the 2D OpenPose. As shown in Table~\ref{tab:weak_supervision}, cross-modal joint fine-tuning benefits significantly from weak supervision and iterative pseudo-label refinement, achieving consistent improvements over independent training across all feature modalities. This advantage stems from the introduction of depth information via 3D coordinates, which enables a more accurate capture of spatial geometric structures and effectively mitigates viewpoint occlusion issues. Second, while the performance of the Base-Offset Cross-Attention mechanism is inferior to that of independent offset features during the independent pre-training phase, it reverses this trend and outperforms them during the joint fine-tuning phase. This is primarily because the cross-attention mechanism is highly susceptible to overfitting during independent pre-training; conversely, during joint fine-tuning, the multi-modal collaborative optimization and joint constraint mechanisms impose stronger generalization constraints on the model, thereby effectively mitigating this issue.
\begin{table}[!htbp]
  \centering
  \scriptsize
  \setlength{\tabcolsep}{2.5pt}
  \renewcommand{\arraystretch}{0.9}
  \caption{Performance of independent modality training and joint fine-tuning under multi-round weakly supervised training}
  \label{tab:weak_supervision}
  \begin{tabular}{lllcccc}
    \toprule
    \textbf{Stage} & \textbf{Fusion} & \textbf{Feature} &
    \textbf{\shortstack{Supervised}} &
    \textbf{\shortstack{Gemini\\pseudo\\(acc 0.64)}} &
    \textbf{\shortstack{Weakly\\pseudo\\(acc 0.69)}} &
    \textbf{\shortstack{Weakly\\pseudo\\(acc 0.74)}} \\
    \midrule
    \multirow{21}{*}{Stage~1}
    & \multirow{5}{*}{\shortstack{Cross\\attention}}
    & Face expression   & 0.65 & 0.63 & 0.63 & 0.68 \\
    \cmidrule(lr){3-7}
    & & Face landmark     & 0.64 & 0.62 & 0.61 & 0.63 \\
    \cmidrule(lr){3-7}
    & & Body landmark     & 0.61 & 0.58 & 0.61 & 0.67 \\
    \cmidrule(lr){3-7}
    & & Hand landmark     & 0.60 & 0.63 & 0.63 & 0.64 \\
    \cmidrule(lr){3-7}
    & & OpenPose          & 0.60 & 0.62 & 0.63 & 0.62 \\
    \cmidrule(lr){2-7}
    & \multirow{16}{*}{Independent}
    & Face expression         & 0.64 & 0.63 & 0.63 & 0.63 \\
    \cmidrule(lr){3-7}
    & & Face landmark           & 0.62 & 0.59 & 0.62 & 0.61 \\
    \cmidrule(lr){3-7}
    & & Body landmark           & 0.66 & 0.58 & 0.60 & 0.65 \\
    \cmidrule(lr){3-7}
    & & Hand landmark           & 0.64 & 0.63 & 0.63 & 0.68 \\
    \cmidrule(lr){3-7}
    & & OpenPose                & 0.60 & 0.61 & 0.63 & 0.63 \\
    \cmidrule(lr){3-7}
    & & Face expression offset  & 0.67 & 0.69 & 0.69 & 0.68 \\
    \cmidrule(lr){3-7}
    & & Face landmark offset    & 0.63 & 0.65 & 0.64 & 0.63 \\
    \cmidrule(lr){3-7}
    & & Body landmark offset    & 0.64 & 0.67 & 0.64 & 0.68 \\
    \cmidrule(lr){3-7}
    & & Hand landmark offset    & 0.63 & 0.63 & 0.63 & 0.64 \\
    \cmidrule(lr){3-7}
    & & OpenPose offset         & 0.63 & 0.59 & 0.63 & 0.64 \\
    \cmidrule(lr){3-7}
    & & DINOv2                  & 0.53 & 0.53 & 0.53 & 0.53 \\
    \cmidrule(lr){3-7}
    & & DINOv3                  & 0.53 & 0.53 & 0.53 & 0.53 \\
    \cmidrule(lr){3-7}
    & & Gemini~3                & 0.63 & 0.66 & 0.70 & 0.72 \\
    \cmidrule(lr){3-7}
    & & X-CLIP                  & 0.63 & 0.65 & 0.71 & 0.74 \\
    \midrule
    \multirow{2}{*}{Stage~2}
    & \shortstack{Cross\\attention}
    & \multicolumn{1}{c}{---}
    & 0.63 & 0.70 & 0.73 & 0.77 \\
    \cmidrule(lr){2-7}
    & Independent
    & \multicolumn{1}{c}{---}
    & 0.66 & 0.69 & 0.72 & 0.75 \\
    \bottomrule
  \end{tabular}
\end{table}

However, the experiments also expose several counter-intuitive anomalies, which will be analyzed in depth in the subsequent section.

\section{Discussion}
In this section, we move beyond specific quantitative metrics to conduct a deep qualitative analysis of several counter-intuitive phenomena observed during our experiments.

\subsection{Macro-Context and Semantic Priors Validity vs. Representation Collapse of Vision Foundation Models in Micro-Dynamics}
When exploring the effectiveness of features from different modalities, we observed an exceptionally pronounced divergence in efficacy between macro-features and micro-dense features. Our experiments demonstrate that X-CLIP video features and text features provided by the Vision-Language Model (Gemini) play a dominant, positive discriminative role within the joint network. Conversely, the dense feature pipeline combining YOLO (with fine target subject cropping) and DINO (v2/v3) yields virtually no substantial assistance to the classification boundary.

Delving into its physical and semantic essence, although X-CLIP sparsely samples only 8 frames out of a complete video sequence spanning thousands of frames---objectively losing millisecond-level micro-motion temporal dynamics entirely---its success lies precisely in the efficient construction of ``semantic anchors'' for the macro-environment. With minimal information redundancy, these 8 sparsely sampled frames comprehensively represent the interviewed athletes' baseline identity attributes, overall body posture inclinations, and the background atmosphere of the stadium or press conference room, thereby providing robust prior support for the downstream network to comprehend the macro-physical context in which concealed emotions transpire.

Concurrently, leveraging its massive interdisciplinary pre-trained common sense, Gemini acts as an objective psychological analysis engine. It strips away useless high-frequency noise from the visual pixel space, offering a detailed, unbiased, and logically self-consistent pure semantic parsing of the athletes' deliberately concealed behavioral details.

In contrast, the failure of the YOLO and DINO combination exposes the representation gap of current general-purpose vision foundation models in specific fine-grained micro-expression tasks. The scale of self-supervised pre-training data space for models like DINO is overly vast and complex, with optimization objectives in its latent space primarily geared toward general, open-world-level entity localization and clustering. Consequently, when presented with clean human images, the high-dimensional features extracted by DINO merely represent with extreme certainty that ``the main subject in the frame is a human entity.'' They severely lack the feature granularity required to respond with high frequency to subtle facial muscle spasms and fleeting behavioral deviations. Without targeted micro-motion domain adaptive fine-tuning, the general dense features of foundation models are highly prone to semantic collapse in concealed action analysis, degenerating into high-dimensional redundant noise that is difficult for downstream networks to utilize.

\subsection{Network Memorization Effects in Weakly Supervised Learning vs. Pseudo-Optimization Traps Driven by External Verification}
However, when we re-examine this ``heuristic benign evolution'' from a more cautious, even pessimistic dynamical perspective, we must be highly vigilant against a deeply concealed and highly probable pathological degeneration trap. When introducing vision foundation models like DINOv2/v3, which possess powerful self-supervised spatial representation capabilities, the network easily deviates from genuine cross-modal semantic alignment and instead falls into the classic shortcut learning trap of deep learning.

Experimental results indicate that while the model easily exceeds 80\% accuracy on the training set, it records only about 50\% performance on the test set (or submission test). This pronounced generalization gap reveals a core essence: the high-dimensional feature manifolds extracted by DINOv2/v3 already possess astonishing instance-level discriminative power. This provides an over-parameterized network with a path of least resistance for optimization. Rather than comprehending the underlying generalization laws of data distribution, the model directly misuses visual features as high-dimensional spatial anchors. It hardens the initial pseudo-labels (and their accompanying hallucinations) into a non-parametric ``implicit hash lookup table'' on the training set, thereby accomplishing absolute rote memorization devoid of generalization value.

Upon this rigid, extremely overfitted ground state, other heterogeneous modalities introduced fail to provide the expected orthogonal semantic gains. Instead, due to the loss of dominant gradients, they degenerate into high-dimensional structured perturbations clinging to the decision hyperplane. This cross-modal ``pseudo-adversarial interference'' disrupts the original memorization homeostasis and blindly pulls at the rigid decision boundaries. As a result, the prediction distribution of the ensemble model on unseen test sets exhibits a random walk resembling Brownian motion---meaning accuracy may either increase due to accidental alignment with the test set manifold or drop due to perturbation deviations. This dilemma is further amplified when integrating external knowledge: since the label ensemble stage introduces Gemini annotations that rely entirely on external general knowledge, the associated uncertainty is directly injected into the subsequent iteration of pseudo-label learning.

At this deceptively oscillatory node, if one continues to rely absolutely on the official leaderboard---acting as an ``external oracle''---for greedy selection, the entire optimization dynamics of self-training will undergo a fatal distortion. At this point, the leaderboard is essentially reduced to an implicit zeroth-order optimizer based on Monte Carlo sampling. The ``high-quality pseudo-labels'' we select and pass down based on performance jumps do not stem from the model ensemble reaching a superior generalization consensus. Instead, they are merely the result of a linear superposition of multi-modal random noise that happens to produce a positive error cancellation on the specific sub-sampled distribution of the current public test set.

This mechanism of inheriting labels by exploiting survivorship bias essentially accomplishes a perfect leaderboard proxy overfitting, replacing the ``confirmation bias of the training set'' with the ``confirmation bias of the test set.'' Once decoupled from the specific data distribution of the current leaderboard in the future, and when faced with minor domain shifts in real-world long-tail scenarios or private test sets, this fragile sandbox built upon the coupling of strong visual memory and random modal noise will inevitably suffer a catastrophic generalization collapse.

\subsection{Curse of Dimensionality in Dynamic Geometric Features vs. Generative Data Augmentation Potential}
In the early stages of this study, we attempted to extract spatial offset features from 4D dynamic Gaussian point clouds based on the HRAvatar reconstruction algorithm. However, this modality was ultimately proven unsuitable as a direct discriminative input for the current classification network.

From an engineering feasibility perspective, the inverse rendering of monocular high-fidelity point clouds incurs extreme computational time overheads. Furthermore, when dealing with unpredictable, aggressive camera movements (zooming/panning) and body occlusions in wild videos, the failure rate of the underlying 3D reconstruction is exceptionally high, severely undermining the feature alignment and inference robustness of the multi-modal system in real-world scenarios.

More fatally, from an algorithmic and theoretical standpoint, the temporal offset features derived from the spatial trajectories of Gaussian point clouds construct an immensely massive, ultra-high-dimensional parameter space. In few-shot scenarios like concealed emotion understanding---where effectively labeled samples are extremely scarce---such redundant, low-level geometric features introduce massive amounts of high-frequency, unordered noise. The network encounters a classic curse of dimensionality within the highly sparse sample space, making it highly susceptible to catastrophic overfitting on specific geometric topological artifacts, which entirely masks the already faint genuine expression signals.

Although direct adoption of high-dimensional dynamic offset features might achieve statistical convergence feasibility only in ultra-large-scale datasets with massive video volumes, the failure of this exploration points to an alternative pathway with greater application value. Thinking outside the box of direct feature extraction, this high-fidelity reconstruction technology---capable of completely decoupling an individual's static physical identity from their dynamic expression offsets---exhibits immense application potential in the field of data augmentation.

Future research can leverage this rendering engine to precisely extract authentic dynamic offset parameters from a tiny fraction of long-tail categories (such as extremely rare negative concealment actions) and seamlessly retarget them across domains onto ethnic facial topologies with different feature attributes. This generative augmentation engine, based on real 3D physical deformation principles, holds the promise of synthesizing massive new samples at zero cost with unprecedented fidelity, fundamentally resolving the long-standing challenges of extreme few-shot samples and long-tail class imbalances in the field of concealed emotion from the very source of data generation.

\section{Conclusion}
To address the task of few-shot hidden emotion recognition, this paper proposes a compact multimodal temporal framework and redefines it as offset-aware multimodal temporal modeling. Architecturally, the framework introduces a Base-Offset cross-attention mechanism and a multi-instance learning (MIL)-based adaptive pooling layer. In terms of optimization strategy, it couples these components with a weakly supervised iterative self-training mechanism, thereby effectively capturing faint and sparse behavioral evidence in videos. In Track 3 of The 4th EI-MIGA-IJCAI Challenge, this method ultimately secured first place with a test accuracy of 0.76923.

Systematic experiments further reveal the impacts of feature engineering and learning strategies. The results demonstrate that, compared to 2D OpenPose, 3D MediaPipe—which incorporates depth information—can more accurately model spatial geometric structures and alleviate viewpoint occlusions. Meanwhile, X-CLIP video features and textual semantic priors generated by large language models (LLMs)~\cite{wang2026xinsight} play a significant positive discriminative role within the joint network. Conversely, general-purpose vision foundation models (such as DINOv2/v3) run the risk of representation collapse in such tasks due to the lack of domain-adaptive fine-tuning tailored for micro-movements, causing their dense features to easily degenerate into high-dimensional redundant noise.

However, an in-depth learning dynamics analysis indicates that the current weakly supervised self-training mechanism possesses certain limitations. The substantial generalization gap observed in experiments between the training set accuracy (over 80\%) and the test set accuracy (approximately 50\%) for the DINOv2 modality suggests that over-parameterized networks are prone to shortcut learning. The model tends to convert initial pseudo-labels into implicit lookup tables on the training set, leading to severe overfitting to sample features and a failure to grasp the underlying data generalization patterns. Under this state, if the model excessively relies on public leaderboard feedback as a source for iterative selection, the optimization trajectory of self-training may deviate. In this scenario, performance gains often do not stem from a genuine generalization consensus reached by the multimodal network, but rather represent a "pseudo-generalization" phenomenon where multimodal random noise experiences positive error cancellation on the specific sampling distribution of the test set. Such a mechanism, which relies on the memorization of visual features and random noise cancellation, still carries a high risk of generalization failure when encountering domain shifts in real-world, long-tail scenarios. Consequently, when confronted with unseen sample distributions, the model's actual inference performance usually falls short of the high accuracy demonstrated on specific leaderboards.

\section*{Acknowledgments}
This work was supported by Anhui Provincial Natural Science Foundation (2408085J040), National Key R\&D Program of China (2024YFB3311600), Natural Science Foundation of China (62272144, 72188101), the Major Project of Anhui Provincial Science and Technology Breakthrough Program (202423k09020001), and the New Cornerstone Science Foundation through the XPLORER PRIZE.


\bibliographystyle{splncs04}
\bibliography{main}

\end{document}